\documentclass{article}
\usepackage{spconf,amsmath,graphicx}
\usepackage{booktabs}
\usepackage{amsmath}

\usepackage{multirow}
\usepackage{spconf}
\usepackage{times}
\usepackage{epsfig}
\usepackage{graphicx}
\usepackage{amsmath}
\usepackage{amssymb}
\usepackage{booktabs}
\usepackage{mathrsfs}
\usepackage{bm}
\usepackage{algorithm, algorithmic}
\usepackage{mathdots}
\usepackage{multirow}
\usepackage[american]{babel}
\usepackage{microtype} 
\usepackage{array}
\usepackage{setspace}
\usepackage{threeparttable}
\usepackage{graphicx}
\usepackage{diagbox}
\usepackage{wrapfig}
\usepackage{float}
\usepackage{bbm}
\usepackage{url}
\usepackage{graphicx}
\usepackage{color}

\title{Selective Domain-Invariant Feature for Generalizable Deepfake Detection}
\name{Yingxin Lai$^{1}$\qquad Guoqing Yang$^{1}$\qquad Yifan He$^{2}$\qquad Zhiming Luo$^{1,3}$\sthanks{Corresponding author: zhiming.luo@xmu.edu.cn} \qquad Shaozi Li$^{1,3}$}
\address{$^{1}$Department of Artificial Intelligence, Xiamen University, Xiamen, China\\
$^{2}$Reconova Technologies, Xiamen, China\\
\small $^{3}$ Fujian Key Laboratory of Big Data Application and Intellectualization for Tea Industry, Wuyi University, WuyiShan, China}
\begin{document}
%
\maketitle
\begin{abstract}
With diverse presentation forgery methods emerging continually, detecting the authenticity of images has drawn growing attention. Although existing methods have achieved impressive accuracy in training dataset detection, they still perform poorly in the unseen domain and suffer from forgery of irrelevant information such as background and identity, affecting generalizability. To solve this problem, we proposed a novel framework Selective Domain-Invariant Feature (SDIF), which reduces the sensitivity to face forgery by fusing content features and styles. Specifically, we first use a Farthest-Point Sampling (FPS) training strategy to construct a task-relevant style sample representation space for fusing with content features. Then, we propose a dynamic feature extraction module to generate features with diverse styles to improve the performance and effectiveness of the feature extractor. Finally,
a domain separation strategy is used to retain domain-related features to help distinguish between real and fake faces.
Both qualitative and quantitative results in existing benchmarks and proposals demonstrate the effectiveness of our approach.

\end{abstract}
\section{Introduction}

Face forgery technologies~\cite{f2f,nt,fs,deepfake} have been rapidly developed in recent years.
The frequent use of these forged images for political framing, fraud, and privacy concerns has raised significant social issues, prompting researchers to seek ways to address these risks.	

Convolutional neural networks (CNNs)~\cite{xception,mesonet} have shown excellent performance in face forgery detection, but they learn only specific patterns of forgery in the training set and fail to achieve satisfactory performance in other domains. These methods use a convolutional backbone to extract image features and then use a classification header to classify the image, tackling forgery detection as a binary classification problem. 

To improve cross-domain performance, several studies have explored how to extract features standard across forgery methods~\cite{rfm,srm}.
Local-relation~\cite{localrelation} by designing a two-stream branch that fuses transformer features with frequency domain features using a cross-attention mechanism. RCCE~\cite{recce} learns the common representations of real faces by reconstructing face images and mines the differences between real and forgery faces through classification. However, they do not separate features irrelevant to forgery and fail to take full advantage of the intrinsic linkages between different forgery methods. Thus, the mixing of irrelevant information in the features may limit generalization.

\begin{figure}[!t]
\includegraphics[width=1\linewidth,height=0.2\textwidth]{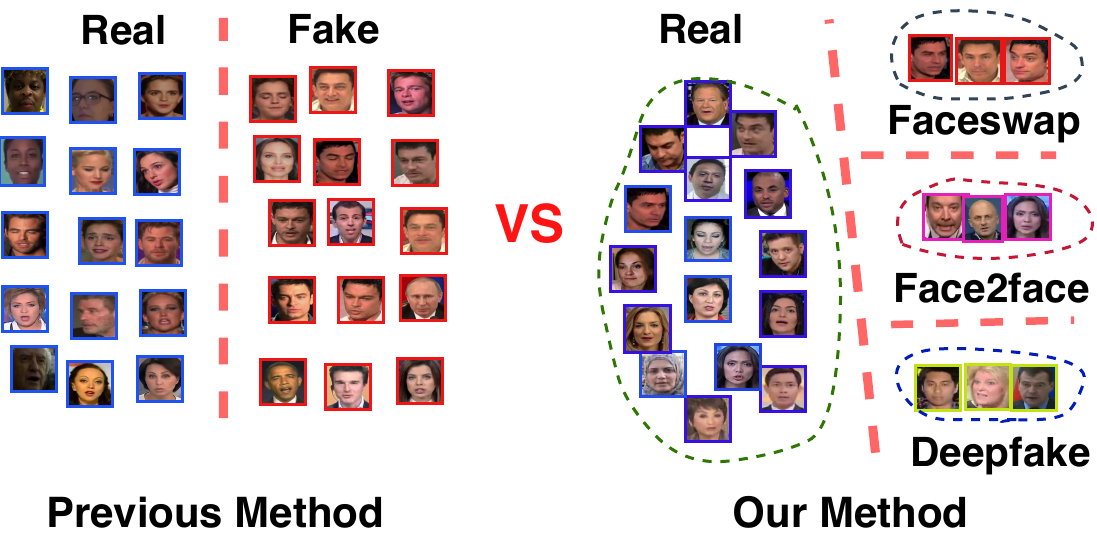}
\caption{Illustration of how our method differs from previous.}
\label{fig:compare_example}
\end{figure}

Domain adaptation (DA) techniques can mitigate the differences between the training and target domains by using domain alignment, yet applying DA directly to the training of deep forgery detection tends to have very unsatisfactory accuracy due to the non-uniformity of the extracted style samples.

Based on the above observations, we propose a novel framework named SDIF, which preserves domain-related features to help distinguish between real and forgery faces. Specifically, we first design an FPS-based style sample sampling strategy by sampling the real and forgery samples separately with FPS, making all the style samples as far away as possible from each other. Thus, the samples are discrete and uniform.Inspired by \cite{zhou2023instance} 
we also propose a Diversity Domain-Aware module (DDA), which fuses content features and style features, embeds domain knowledge in a high-dimensional feature-aware task and then preserves domain knowledge for constraints in the task.

\begin{figure*}[!t]
\centering
\includegraphics[width=\linewidth]{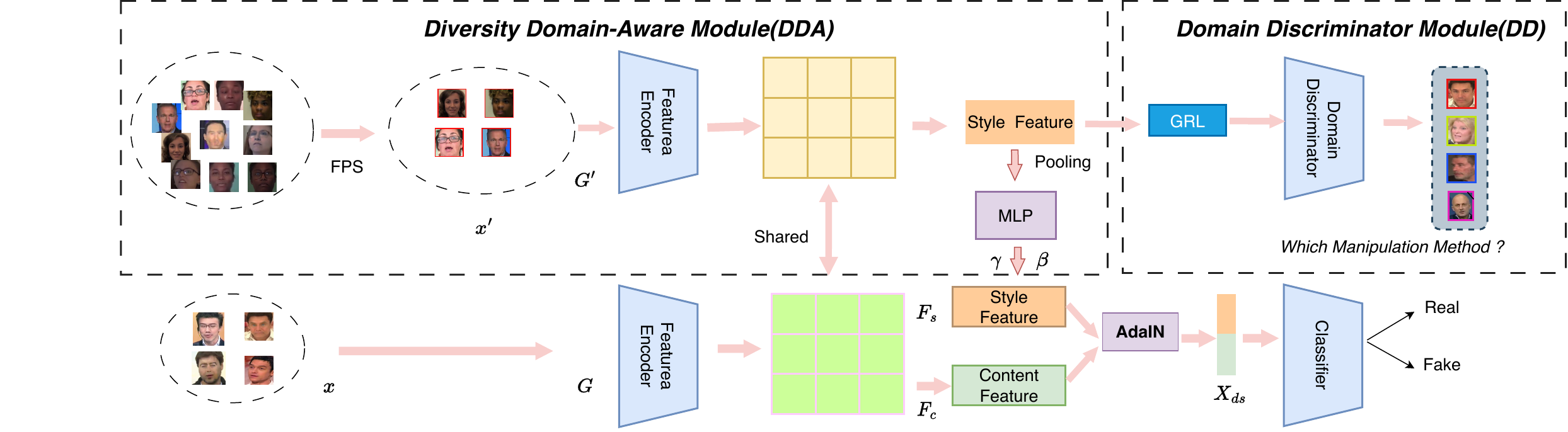}
\caption{The overall framework of our proposed method.}
\label{fig:label_example}
\end{figure*}

\section{METHOD}
The overall framework of our method is presented in Fig ~\ref{fig:label_example}, which contains three major key modules: style sample generation, dynamic feature extractor, and feature fusion domain identification module.

\subsection{Diversity Domain-Aware Module}
In this module, the input $x$ is first fed into the feature extractor $G$ and performing FPS sampling of style samples $x'$ into $G'$ to obtain the feature embedding. The content features and the arbitrary statistic-modified style features are then blended using Adaptive Instance Normalization (AdaIN) to be fed into the domain invariant discriminator $D$, allowing $D$ to concentrate on the input domain invariant features.
 
To maximize the coverage of rare styles as well as to ensure the diversity of sample space inspired by the point cloud sampling algorithms (point cloud sampling algorithms~\cite{pointnet++}), we use FPS to sample the style samples according to the proportion of different categories. In detail, we iteratively select $C$ points from all the points in the $N$ categories to prevent real and fake images from interacting with each other. The point selected in each iteration is the farthest point from the remaining points. After sampling, we use the feature extractor to get feature embeddings. Specifically, we first use a feature generator to capture multi-scale low-level image information. Then, for content features, we obtain $F_{c}$ by batch normalization (BN), while for style features, we use instance normalization (IN) to obtain $S$. Subsequently, we utilize Global Average Pooling (GAP) and Multilayer Perceptron (MLP) to derive $F_{s}$, which is employed for the fusion of content information.
\begin{equation}
\begin{split}
F_{s}= \text{MLP}\left(\text{GAP}(S)\right)
\end{split}
\end{equation}


AdaIN is an adaptive stylization module; given a content input $A$ and a randomly chosen input style input $A'$, the stylized normalization can be formed as follows:
\begin{equation}
\begin{split}
\text{AdaIN}(A,A',\mu,\sigma)=\gamma\left( A' \right)\left( \frac{A  - \mu\left( A \right)}{\sigma\left( A \right)} \right)+\beta\left( A' \right)
\end{split}
\end{equation}
where $\mu\left(\cdot\right)$ and $\sigma\left(\cdot\right)$ denotes the channel-wise mean and standard deviation.$\gamma\left(\cdot\right)$ and $\eta\left(\cdot\right)$ are the affine parameters generated by the style input $A'$.
For each category N,we sample the combination weight $W^{n}=\left[ w^{1}...,w^{c} \right]$ 
from Dirichlet distribution 
$B\left(\left[\alpha^{1}...,\alpha^{c} \right] \right)$
with the concentration parameters $\left[ \alpha^{1} ...,\alpha^{c} \right]$ 
all set to $1/C$.
Then, the basis styles of each class $N$ are linearly combined by $W^{n}$ for the aggregation.
\begin{equation}
\begin{split}
\mu_{F_{s}}=W^{n}\cdot \mu_{base},\quad    \sigma_{F_{s}}=W^{n}\cdot \sigma_{base}
\end{split}
\end{equation}
where $\mu_{base}$ and $\sigma_{base}$ are the C bias styles.

With the generated styles, diversity style $X_{ds}$ are:
\begin{equation}
\begin{split}
X_{ds}= \gamma^{n}\left( F_{s} \right)\left( \frac{F_{c}-\mu\left( F_{c} \right)}{\sigma\left( F_{c} \right)} \right)+\beta^{n}\left( F_{s} \right)
\end{split}
\end{equation}

where $\mu\left(\cdot\right)$ and $\sigma\left(\cdot\right)$ denotes the channel-wise mean and standard deviation.$\gamma\left(\cdot\right)$ and $\beta\left(\cdot\right)$ are the affine parameters generated by the style input $F_{s}$.

\subsection{Dynamic Feature Extractor}

The utilization of static convolution poses certain challenges in extracting adaptive features, primarily stemming from the diverse array of forgery methods. To address this issue, inspired by \cite{zhou2023instance}, we have introduced a Dynamic Feature Extraction module(DFE). As shown in Fig ~\ref{fig: DFE}, One part is fed into the dynamic convolutional branch, and the other part is passed to the static convolutional branch. The dynamic convolutional branch acquires the representation of half of the channels. This module is added after the convolutional layer of each feature extractor. The input feature $M$ is split into two parts $M_{a}$  and $M_{b}$ after channel split. 
\begin{equation}
\begin{split}
Z=\nu\left( M_{a} \right)\otimes M_{a} ,\quad Z'=Conv\left( M_{b} \right)
\end{split}
\end{equation}
where $\nu\left( \cdot  \right)$ denotes dynamic convolution operation, $Conv$ is a convolution block ,in which $\nu$ depends on the input instance $M_{a}$ and $\otimes$ denotes the element-wise multiplication.We then concatenate $M_{a}$ and $M_{b}$ in channel dimension and feed the result into a convolution block $\delta\left( \cdot  \right)$ to produce output features $F$, denoted as:
\begin{equation}
\begin{split}
 F=\delta\left( concate(Z,Z') \right)
\end{split}
\end{equation}

\begin{figure}[!t]
\centering
\includegraphics[width=1\linewidth]{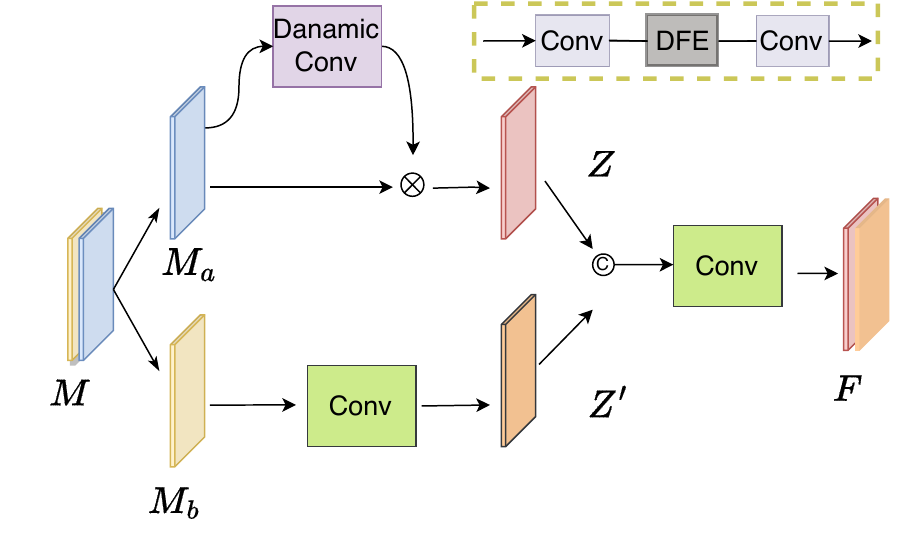}
\caption{Illustrations of our designed Dynamic feature Extractor.}
\label{fig:DFE}
\end{figure}

\subsection{Domain Discriminator Module}
Relevant research~\cite{control} has shown evidence that it is crucial to establish an appropriate level of differentiation among different forged methods. At the same time, we also want to make more general content features, this process can be learned through the following optimization problem:

\begin{equation}
\begin{aligned}
& \min _D \max _G L_{a d v}(G, D)= \\
& -\mathbb{E}_{(x, y) \sim\left(X, Y_N\right)} \sum_{i=1}^N [i=y] \log D(G(x)),
\end{aligned}
\end{equation}
where $N$ denotes the number of domains,  $x$ represents the training set of samples, and $y$ is the set of binary labels. $Y_{n}$ corresponds to the one-hot encoding of the domain label.$G$ and $D$ denote the content feature generator and domain discriminator, respectively. To simultaneously optimize $G$ and $D$,the gradient reversal layer (GRL)~\cite{grl} is employed by us to reverse the gradient during the backward process.

\subsection{Loss Function}
The total loss function $L$ of the proposed framework includes the domain differentiation loss for domain learning and  the cross-entropy loss $L_{ce}$ for binary classification, which can be presented as follows:
\begin{equation}
\begin{aligned}
L=L_{ce}+\lambda L_{adv},
\end{aligned}
\end{equation}
where $\lambda$ is the hyperparameter for balancing the two losses.
\section{EXPERIMENT}

\subsection{Experimental Setting}
\textbf{Datasets.} Following previous studies~\cite{instance,DGfas}, we validated our method using four publicly available datasets: FF++~\cite{ff++}, Celeb-DF~\cite{celebdf}, WildDeepfake~\cite{wildfake}, and DFDC ~\cite{dfdc}. \par

\textbf{Implementation details.} 
Our experiments were conducted using PyTorch on a single RTX-3090 GPU. For training, we utilized ResNet18  pre-trained on ImageNet.
The batch size is set to 32, and Adam is used for optimization with an initial learning rate of 2e-5. Additionally, we used RetinaFace for face extraction and alignment, resizing the aligned faces to 229×229. The implementation of dynamic convolution was based on ~\cite{danamicconv}.\par
\textbf{Evaluation Metrics.} 
For evaluation, we utilize metrics commonly employed in related research~\cite{rfm,recce,f3net,facexray,fwa}, including ACC, AUC, and EER.

\subsection{Comparison with Other Methods}
In this section, we compare the performance of our method with those of the current state-of-the-art model~\cite{recce,rfm, spsl,localrelation}.
We first assess the performance within the FF++~\cite{ff++} datasets. Following that, we examine cross-dataset generalization by training our model on FF++(HQ)~\cite{ff++} and evaluating its performance on Celeb-DF~\cite{celebdf}, DFDC~\cite{dfdc}, and FF++(LQ)~\cite{ff++} datasets, it was additionally trained using the NT forgery mode and subsequently evaluated on several different forgery types.
\begin{table}[t]
\renewcommand{\arraystretch}{1.2} 
\centering 
\caption{In-domain comparisons on FF++ dataset. Results contain ACC(\%) and AUC (\%)  of high quality (HQ) and low quality (LQ).}
\label{tab:indomain}
\resizebox{0.48\textwidth}{29.5mm}{%
\begin{tabular}{l|cc|cc}
\hline
\multicolumn{1}{c|}{\multirow{2}{*}{Methods}} & \multicolumn{2}{c|}{FF++ (HQ)} & \multicolumn{2}{c}{FF++ (LQ)} \\
\cline{2-5}
& ACC & AUC & ACC & AUC \\
\hline
Xception ~\cite{xception} & 95.04 & 96.30 & 84.11 & 92.50 \\
F$^3$-Net (Xception) ~\cite{f3net} & 97.31 & 98.10 & 86.89 & 93.30 \\
EN-B4 ~\cite{efb4} & 96.63 & 99.18 & 86.67 & 88.20 \\
MAT (EN-B4) ~\cite{mat} & 97.60 & 99.29 & 88.69 & 90.40 \\
SPSL ~\cite{spsl} & 91.50 & 95.32 & 81.57 & 82.82 \\
RFM ~\cite{rfm} & 95.69 & 98.79 & 87.06 & 89.83 \\
Local-relation ~\cite{localrelation} & 97.59 & 99.56 & 91.47 & 95.21 \\
RECCE ~\cite{recce} & 97.06 & 99.32 & 91.03 & 95.02 \\
\hline
\textbf{Ours} & \textbf{99.19} & \textbf{99.93} & \textbf{92.17} & \textbf{96.14} \\
\hline
\end{tabular}
}
\vspace{0.01cm}
\vspace{-0.2cm}
\end{table}

\begin{table}[t]
\caption{Cross-domain comparisons of generalization based on AUC (\%). We train the model on the HQ dataset of FF++ and evaluate it on Celeb-DF and DFDC.}\label{tab:crossdomain}
\centering 
\begin{tabular}{l|ccc}
\hline
Methods                           & Celeb-DF        & DFDC          \\ \hline
Xception ~\cite{xception}               & 66.91          & 67.93         \\
F$^3$-Net (Xception) ~\cite{f3net}      & 71.21          & 72.88          \\
EN-B4 ~\cite{efb4}                     & 66.24          & 66.81          \\
MAT(EN-B4) ~\cite{mat}                 & 76.65          & 67.34         \\
Face X-ray ~\cite{spsl}                & 74.20          & 70.00          \\
RFM ~\cite{rfm}                       & 67.64          & 68.01       \\
SRM ~\cite{srm}                         & 79.40          & 79.70         \\
Local-relation ~\cite{localrelation}    & 78.26          & 76.53          \\
RECCE ~\cite{recce}                    & 77.39          & 76.75         \\
LTW ~\cite{ltw}                       & 77.14          & 74.58        
   \\ \hline
\textbf{Ours}                         & \textbf{81.79} & \textbf{80.65} \\ 
\hline
\end{tabular}
\vspace{0.05cm}
\vspace{-0.18cm}
\end{table}

\textbf{Intra-dataset Testing.}
In Table~\ref{tab:indomain}, we present our findings on intra-domain detection performance. Our accuracy scores (ACCs) on FF++(LQ) and Celeb-DF are impressive at 99.19\% and 92.17\%, respectively, surpassing the results of other methods listed in the table. For comparison, the recent RECCE network achieves ACCs that are 2.13\% and 1.14\% lower than ours. Although RECCE demonstrates strong performance in accurately learning the distribution of real faces through reconstruction, reaching impressive AUC scores of 99.32\% and 95.02\% on the corresponding datasets, it still lags behind our technique by an average margin of 0.86\%. This highlights the effectiveness of our method, which focuses on different forgery categories, allowing us to capture more forgery-related information.

\textbf{Cross-dataset Testing.}
To validate the generalization performance of our method, we trained it on FF++HQ data and then tested it in other domains. Table ~\ref{tab:crossdomain} evident that our method achieves the best performance on DFDC and Celeb-DF, with AUC scores of 81.79\% and 80.65\%, respectively.
In comparison, the SRM model ~\cite{srm}, which integrates RGB and frequency-domain information, performs reasonably well with AUC scores of 79.40\% and 74.70\%, respectively. However, its performance remains inferior to our method. Our method shows superior performance on undetected forgery types compared to previous methods, as shown in table ~\ref{cross_domain_nt}. These results confirm the feasibility of using domain separation and improved stylistic features to detect forgeries with unknown patterns.

\begin{table}[]
\centering
\caption{Cross-database evaluation on FF++ database (HQ).}
\label{cross_domain_nt}
\scalebox{0.85}{

\begin{tabular}{c|c|lll}
\hline
\multirow{2}{*}{Training Set} & \multirow{2}{*}{Method} & \multicolumn{3}{c}{Testing Set(AUC)} \\ \cline{3-5} 
                              &                         &  \multicolumn{1}{c}{DF}     &  \multicolumn{1}{c}{FS}    & \multicolumn{1}{c}{F2F}   \\ \hline
\multirow{6}{*}{NT}           & Xception ~\cite{xception}                & 76.98      & 70.73      & 73.57      \\ \cline{2-2}
                              & Face X-ray ~\cite{facexray}              & 65.31      & 74.68      & 74.86      \\ \cline{2-2}
                              & F$^3$-Net (Xception) ~\cite{f3net}    & 81.50      & 64.49      & 77.18      \\ \cline{2-2}
                              & RFM ~\cite{rfm}                    & 80.05      & 64.93      & 74.50      \\ \cline{2-2}
                              & RECCE ~\cite{recce}                  & 78.83      & 63.70      & 80.89      \\ \cline{2-5} 
                              & \textbf{Our}            &  \textbf{83.54}          & \textbf{75.14}           &  \textbf{83.69}          \\ \hline
\end{tabular}
}
\end{table}

\vspace{-0.6cm}
\subsection{Ablation Studies and Visualization}
\begin{table}[]
\centering
\caption{Ablation study on  the proposed components in our method. Cross-dataset evaluation on WildDeepfake dataset.}
\label{abalation_module}
\begin{tabular}{llll|cc}
\hline
\multicolumn{1}{c}{Baseline} & \multicolumn{1}{c}{DDA} & \multicolumn{1}{c}{DFE} & \multicolumn{1}{c|}{DD} & AUC & EER \\ \hline
\checkmark                   &                         &                         &                         & 61.21   & 45.67   \\
\checkmark                   & \checkmark              &                         &                         & 62.76   & 43.72  \\
\checkmark                   & \checkmark              & \checkmark              &                         & 63.22   & 43.04  \\
\checkmark                   & \checkmark              & \checkmark              & \checkmark              & \textbf{64.01}  & \textbf{41.54}  \\ \hline
\end{tabular}
\end{table}

\textbf{Importance of different modules.}
The key components of the performance of our method are shown in Table ~\ref{abalation_module}, where the model is trained on FF++ and tested on WildDeepfake. 
We progressively integrate the DDA, DFE, and DD modules into the baseline model, leading to a gradual enhancement in the performance of the model.
The greatest impact is the addition of the DDA module, e.g., the AUC increases from 61.21\% to 62.76\%. It indicates that the performance of the in-domain representation of diversity is significantly improved, while the ACC and ERR changed by 0.79\% and 1.49\%, respectively, indicating that the DD module contributes to the domain generalization ability of the model.

\textbf{Parameter sensitivity analysis.} In this section, we investigate the effect of balance weight $\lambda$ on the generalization performance, and we retrain the model using different weight coefficients; as can be seen, the overall best performance of ACC is obtained at $\lambda$ equals 1.
\begin{figure}[!t]
\centering
\includegraphics[width=0.8\linewidth]{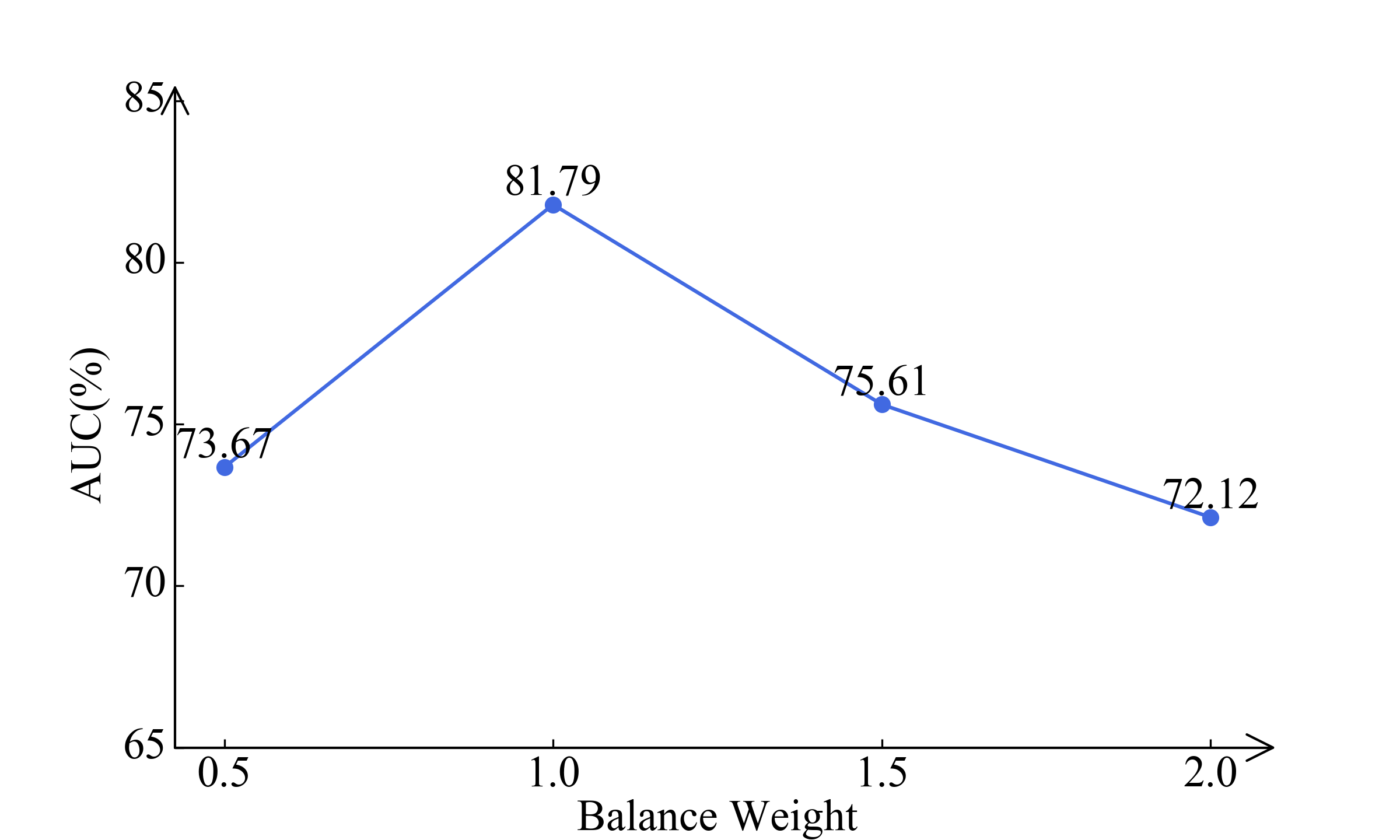}
\caption{Comparative experiments with different balance weight of $\lambda$ from FF++(HQ) to Celeb-DF.}
\label{parameter}
\end{figure}



\section{CONCLUSION}
This paper presents a novel framework to enhance face forgery detection performance in cross-domain. We employ FPS for diverse representations and utilize dynamic convolution to generate varied samples. Additionally, we decouple domain labels to reduce interference from intradomain correlations unrelated to forgery. The effectiveness of our method is demonstrated across multiple benchmark datasets.

{\footnotesize
\noindent\textbf{Acknowledgement:} This work is supported by the National Natural Science Foundation of China (No.~62276221, No.~62376232), and the Science and Technology Plan Project of Xiamen (No. 3502Z20221025), the Open Project Program of Fujian Key Laboratory of Big Data Application and Intellectualization for Tea Industry, Wuyi University (No.~FKLBDAITI202304).
}

\ninept
\bibliographystyle{IEEEbib}
\bibliography{strings,refs}

\begin{thebibliography}{10}

\bibitem{f2f}
Justus Thies, Michael Zollhofer, Marc Stamminger, Christian Theobalt, and
  Matthias Nie{\ss}ner,
\newblock ``Face2face: Real-time face capture and reenactment of rgb videos,''
\newblock in {\em Proceedings of the IEEE conference on computer vision and
  pattern recognition}, 2016, pp. 2387--2395.

\bibitem{nt}
Justus Thies, Michael Zollh{\"o}fer, and Matthias Nie{\ss}ner,
\newblock ``Deferred neural rendering: Image synthesis using neural textures,''
\newblock {\em ACM Transactions on Graphics (TOG)}, vol. 38, no. 4, pp. 1--12,
  2019.

\bibitem{fs}
``Faceswap,'' \url{https://github.com/MarekKowalski/FaceSwap},
\newblock Accessed 2022-10-29.

\bibitem{deepfake}
``Deepfakes,'' \url{https://github.com/deepfakes/faceswap},
\newblock Accessed 2022-10-29.

\bibitem{xception}
Andreas Rossler, Davide Cozzolino, Luisa Verdoliva, Christian Riess, Justus
  Thies, and Matthias Nie{\ss}ner,
\newblock ``Faceforensics++: Learning to detect manipulated facial images,''
\newblock in {\em Proceedings of the IEEE/CVF international conference on
  computer vision}, 2019, pp. 1--11.

\bibitem{mesonet}
Darius Afchar, Vincent Nozick, Junichi Yamagishi, and Isao Echizen,
\newblock ``Mesonet: a compact facial video forgery detection network,''
\newblock in {\em IEEE international workshop on information forensics and
  security (WIFS)}. IEEE, 2018, pp. 1--7.

\bibitem{rfm}
Chengrui Wang and Weihong Deng,
\newblock ``Representative forgery mining for fake face detection,''
\newblock in {\em Proceedings of the IEEE/CVF conference on computer vision and
  pattern recognition}, 2021, pp. 14923--14932.

\bibitem{srm}
Yuchen Luo, Yong Zhang, Junchi Yan, and Wei Liu,
\newblock ``Generalizing face forgery detection with high-frequency features,''
\newblock in {\em Proceedings of the IEEE/CVF conference on computer vision and
  pattern recognition}, 2021, pp. 16317--16326.

\bibitem{localrelation}
Shen Chen, Taiping Yao, Yang Chen, Shouhong Ding, Jilin Li, and Rongrong Ji,
\newblock ``Local relation learning for face forgery detection,''
\newblock in {\em Proceedings of the AAAI Conference on Artificial
  Intelligence}, 2021, vol.~35, pp. 1081--1088.

\bibitem{recce}
Junyi Cao, Chao Ma, Taiping Yao, Shen Chen, Shouhong Ding, and Xiaokang Yang,
\newblock ``End-to-end reconstruction-classification learning for face forgery
  detection,''
\newblock in {\em Proceedings of the IEEE/CVF Conference on Computer Vision and
  Pattern Recognition}, 2022, pp. 4113--4122.

\bibitem{zhou2023instance}
Qianyu Zhou, Ke-Yue Zhang, Taiping Yao, Xuequan Lu, Ran Yi, Shouhong Ding, and
  Lizhuang Ma,
\newblock ``Instance-aware domain generalization for face anti-spoofing,''
\newblock in {\em Proceedings of the IEEE/CVF Conference on Computer Vision and
  Pattern Recognition}, 2023, pp. 20453--20463.

\bibitem{pointnet++}
Charles~Ruizhongtai Qi, Li~Yi, Hao Su, and Leonidas~J Guibas,
\newblock ``Pointnet++: Deep hierarchical feature learning on point sets in a
  metric space,''
\newblock {\em Advances in neural information processing systems}, vol. 30,
  2017.

\bibitem{control}
Ying Guo, Cheng Zhen, and Pengfei Yan,
\newblock ``Controllable guide-space for generalizable face forgery
  detection,''
\newblock {\em arXiv preprint arXiv:2307.14039}, 2023.

\bibitem{grl}
Yaroslav Ganin and Victor Lempitsky,
\newblock ``Unsupervised domain adaptation by backpropagation,''
\newblock in {\em International conference on machine learning}. PMLR, 2015,
  pp. 1180--1189.

\bibitem{instance}
Qianyu Zhou, Ke-Yue Zhang, Taiping Yao, Xuequan Lu, Ran Yi, Shouhong Ding, and
  Lizhuang Ma,
\newblock ``Instance-aware domain generalization for face anti-spoofing,''
\newblock in {\em Proceedings of the IEEE/CVF Conference on Computer Vision and
  Pattern Recognition}, 2023, pp. 20453--20463.

\bibitem{DGfas}
Zhuo Wang, Zezheng Wang, Zitong Yu, Weihong Deng, Jiahong Li, Tingting Gao, and
  Zhongyuan Wang,
\newblock ``Domain generalization via shuffled style assembly for face
  anti-spoofing,''
\newblock in {\em Proceedings of the IEEE/CVF Conference on Computer Vision and
  Pattern Recognition}, 2022, pp. 4123--4133.

\bibitem{ff++}
Andreas Rossler, Davide Cozzolino, Luisa Verdoliva, Christian Riess, Justus
  Thies, and Matthias Nie{\ss}ner,
\newblock ``Faceforensics++: Learning to detect manipulated facial images,''
\newblock in {\em Proceedings of the IEEE/CVF international conference on
  computer vision}, 2019, pp. 1--11.

\bibitem{celebdf}
Yuezun Li, Xin Yang, Pu~Sun, Honggang Qi, and Siwei Lyu,
\newblock ``Celeb-df: A large-scale challenging dataset for deepfake
  forensics,''
\newblock in {\em Proceedings of the IEEE/CVF conference on computer vision and
  pattern recognition}, 2020, pp. 3207--3216.

\bibitem{wildfake}
Bojia Zi, Minghao Chang, Jingjing Chen, Xingjun Ma, and Yu-Gang Jiang,
\newblock ``Wilddeepfake: A challenging real-world dataset for deepfake
  detection,''
\newblock in {\em Proceedings of the 28th ACM international conference on
  multimedia}, 2020, pp. 2382--2390.

\bibitem{dfdc}
Brian Dolhansky, Joanna Bitton, Ben Pflaum, Jikuo Lu, Russ Howes, Menglin Wang,
  and Cristian~Canton Ferrer,
\newblock ``The deepfake detection challenge (dfdc) dataset,''
\newblock {\em arXiv preprint arXiv:2006.07397}, 2020.

\bibitem{danamicconv}
Yinpeng Chen, Xiyang Dai, Mengchen Liu, Dongdong Chen, Lu~Yuan, and Zicheng
  Liu,
\newblock ``Dynamic convolution: Attention over convolution kernels,''
\newblock in {\em Proceedings of the IEEE/CVF conference on computer vision and
  pattern recognition}, 2020, pp. 11030--11039.

\bibitem{f3net}
Yuyang Qian, Guojun Yin, Lu~Sheng, Zixuan Chen, and Jing Shao,
\newblock ``Thinking in frequency: Face forgery detection by mining
  frequency-aware clues,''
\newblock in {\em European conference on computer vision}. Springer, 2020, pp.
  86--103.

\bibitem{facexray}
Lingzhi Li, Jianmin Bao, Ting Zhang, Hao Yang, Dong Chen, Fang Wen, and Baining
  Guo,
\newblock ``Face x-ray for more general face forgery detection,''
\newblock in {\em Proceedings of the IEEE/CVF conference on computer vision and
  pattern recognition}, 2020, pp. 5001--5010.

\bibitem{fwa}
Yuezun Li and Siwei Lyu,
\newblock ``Exposing deepfake videos by detecting face warping artifacts.,''
\newblock in {\em CVPR Workshops}, 2019.

\bibitem{spsl}
Honggu Liu, Xiaodan Li, Wenbo Zhou, Yuefeng Chen, Yuan He, Hui Xue, Weiming
  Zhang, and Nenghai Yu,
\newblock ``Spatial-phase shallow learning: rethinking face forgery detection
  in frequency domain,''
\newblock in {\em Proceedings of the IEEE/CVF conference on computer vision and
  pattern recognition}, 2021, pp. 772--781.

\bibitem{efb4}
Mingxing Tan and Quoc Le,
\newblock ``Efficientnet: Rethinking model scaling for convolutional neural
  networks,''
\newblock in {\em International conference on machine learning}. PMLR, 2019,
  pp. 6105--6114.

\bibitem{mat}
Hanqing Zhao, Wenbo Zhou, Dongdong Chen, Tianyi Wei, Weiming Zhang, and Nenghai
  Yu,
\newblock ``Multi-attentional deepfake detection,''
\newblock in {\em Proceedings of the IEEE/CVF conference on computer vision and
  pattern recognition}, 2021, pp. 2185--2194.

\bibitem{ltw}
Ke~Sun, Hong Liu, Qixiang Ye, Yue Gao, Jianzhuang Liu, Ling Shao, and Rongrong
  Ji,
\newblock ``Domain general face forgery detection by learning to weight,''
\newblock in {\em Proceedings of the AAAI conference on artificial
  intelligence}, 2021, vol.~35, pp. 2638--2646.

\end{thebibliography}

\end{document}